\documentclass[letterpaper, 10 pt, conference]{ieeeconf}  

\IEEEoverridecommandlockouts                              

\usepackage{graphicx}
\graphicspath{{figs/}}
\usepackage{amsmath}
\usepackage{amssymb}
\usepackage{algorithm}
\usepackage{algpseudocode}
\usepackage{times}
\usepackage{bm}
\usepackage{color}
\usepackage{mathtools}
\usepackage{algorithm}
\usepackage{algpseudocode}
\usepackage{multirow}
\usepackage{hyperref}
\newcommand{\figref}[1]{Fig.~\ref{#1}}
\newcommand{\tabref}[1]{Table~\ref{#1}}
\newcommand{\secref}[1]{Section~\ref{#1}}
\newcommand{\algoref}[1]{Algorithm~\ref{#1}}
\newcommand{\R}{\mathbb{R}}

\newcommand{\SO}{\mathop{\mathrm{SO}}}
\newcommand{\Ortho}{{\mathrm{O}}}
\newcommand{\ortho}{{\mathfrak{o}}}
\newcommand{\erfc}{\mathop{\mathrm{erfc}}}
\renewcommand{\Re}{\mathop{\mathrm{Re}}}

\newcommand{\TODO}[1]{\textcolor{red}{\textbf{TODO}: #1}}
\renewcommand{\TODO}[1]{\relax}

\newcommand{\D}{{D}}
\newcommand{\eigvals}{\bm{\lambda}}
\newcommand{\normconst}{\mathcal{C}}
\newcommand{\integrand}{\mathcal{F}}
\newcommand{\triu}{\mathop{\mathrm{triu}}}
\newcommand{\diag}{\mathop{\mathrm{diag}}}

\newcommand{\cayley}{\operatorname{cay}}
\newcommand{\birdal}{\operatorname{bir}}

\newcommand{\OmegaL}{\mathop{\mathrm{\Omega_L}}}
\newcommand{\OmegaR}{\mathop{\mathrm{\Omega_R}}}
\newcommand{\qt}{\bm{q}}
\newcommand{\conj}[1]{\overline{#1}}
\newcommand{\qtc}{\conj{\bm{q}}}
\newcommand{\Sp}{\mathbb{S}}
\newcommand{\bingham}{\mathfrak{B}}
\newcommand{\Sym}{\mathrm{Sym}}

\newcommand{\Expect}[1]{\operatorname{E}[#1]}
\newcommand{\trace}{\operatorname{tr}}

\newcommand{\etal}{\textit{et al.}}

\newcommand{\Param}{\mathcal{P}}
\renewcommand{\d}{\mathrm{d}}

\newcommand{\DeltaQ}{\Delta Q}

\title{\LARGE \bf
Probabilistic Rotation Representation With an Efficiently Computable Bingham Loss Function and Its Application to Pose Estimation$^\text{*}$
}

\author{Hiroya Sato$^\text{1,\,2}$, Takuya Ikeda$^\text{1}$, and Koichi Nishiwaki$^{1}$
\thanks{This work has been submitted to the IEEE for possible publication. Copyright may be transferred without notice, after which this version may no longer be accessible.}%
\thanks{$^{\text{*}}$ This is a contribution in Woven Planet Holdings, Inc.
Part of the work is the result of Summer Internship Program.}%
\thanks{$^{\text{1}}$ All authors are with the Woven Planet Holdings, Inc. 3 Chome-2-1 Nihonbashimuromachi, Chuo City, Tokyo, 103-0022, Japan, {\tt\footnotesize [firstname.lastname]@woven-planet.global}}%
\thanks{$^{\text{2}}$Hiroya Sato is with Department of Mechano-Informatics, Graduate School
of Information Science and Technology, The University of Tokyo, 7-3-1
Hongo, Bunkyo-ku, Tokyo, 113-8656, Japan.
{\tt\footnotesize h-sato@jsk.t.u-tokyo.ac.jp}
}%
}%

\begin{document}

\maketitle
\thispagestyle{empty}
\pagestyle{empty}

\begin{abstract}
In recent years, a deep learning framework has been widely used for object pose estimation. 
While quaternion is a common choice for rotation representation of 6D pose, it cannot represent an uncertainty of the observation.
In order to handle the uncertainty, Bingham distribution is one promising solution
because this has suitable features, such as a smooth representation over $\SO(3)$, in addition to the ambiguity representation. However it requires the complex computation of the normalizing constants. This is the bottleneck of loss computation in training neural networks based on Bingham representation.
As such, we propose a fast-computable and easy-to-implement loss function for Bingham distribution. We also show not only to examine the parametrization of Bingham distribution but also an application based on our loss function.

\end{abstract}

\section{INTRODUCTION}

Recently, there are many research efforts on pose estimation based on deep learning framework, such as \cite{xiang2018posecnn, Bui2018-mr}.
In these works, quaternion is widely used for rotation representation. 
However, since single quaternion can only represent a single rotation, it cannot capture an uncertainty of the observation.
Handling uncertainty is quite important, especially in the situation that a target object is occluded or has a symmetric shape \cite{manhardt2019explaining, Hashimoto2019} .

Many researchers have been considering how to represent the ambiguity of rotations. 
One way to represent it is to utilize \textit{Bingham distribution} \cite{bingham1974}.
It mainly has two advantages.
Firstly, the Bingham distribution is the probability distribution that is consistent with the spatial rotation $\SO(3)$ (detaily described in \secref{section:bingham}), and is easy to be parametrized.
Secondly, the continuous representation, which is suitable for a neural network, can be derived from this distribution, as Peretroukhin \etal \cite{peretroukhin_so3_2020} suggested one example of it.
For the above characteristics, we choose the Bingham distribution for probabilistic rotation representation.

To optimize the probability distribution, in general, a negative log-likelihood (NLL) is common choice for loss function.
Since NLL is defined in mathematically natural way, there are plenty of theorems on it. 
Moreover, it has potential of wide application because it can be easily extended to more complicated model, such as mixture model \cite{deng2020deep, Riedel2016}.

\begin{figure}[t]
    \centering
    \includegraphics[width=.8\linewidth]{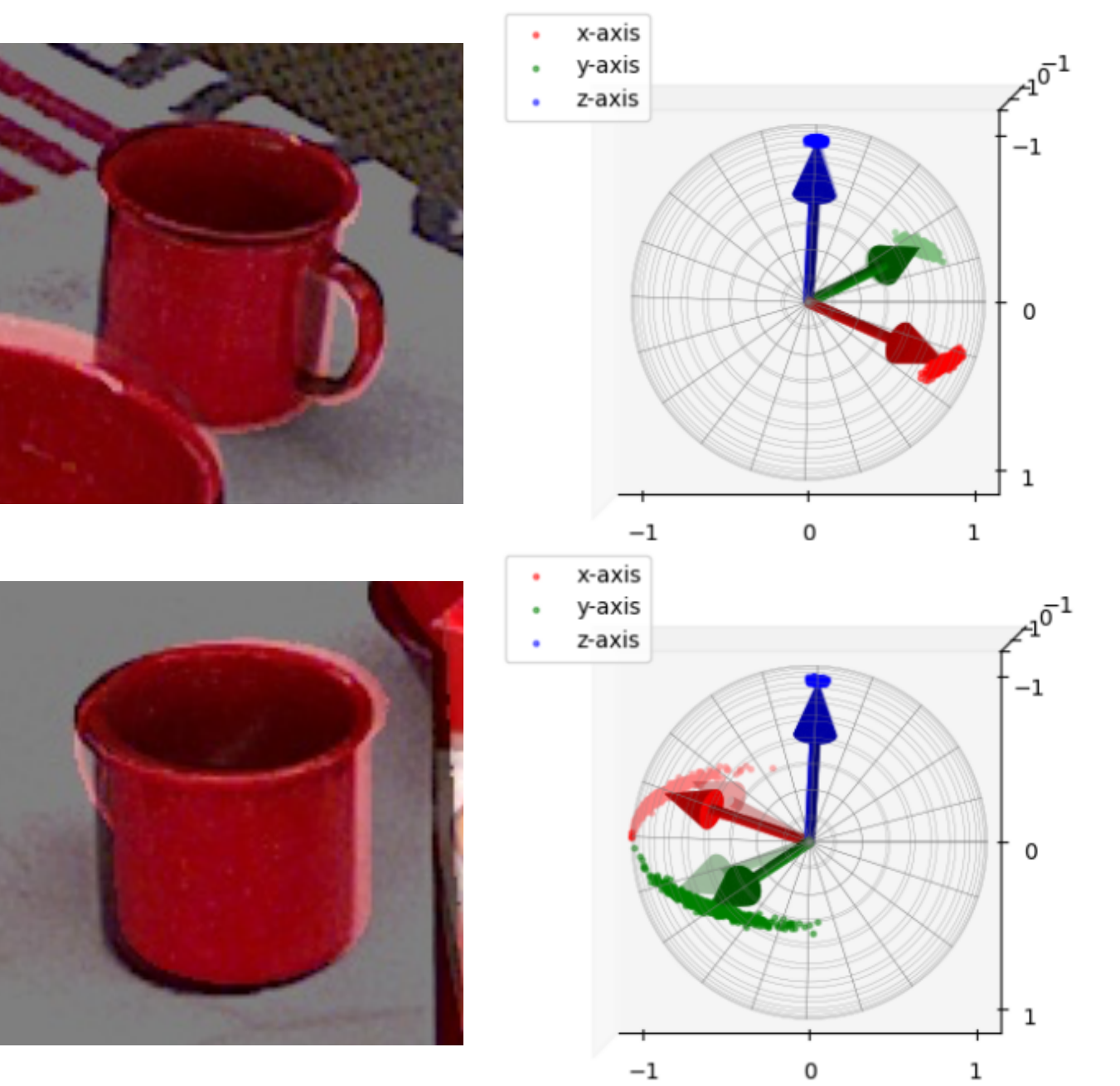}
    \caption{Inference sample of our Bingham model. The upper figure shows that the handle is visible and the corresponding distribution is low-variance, while the lower figure shows that the mug's handle is occluded and the distribution is widely spread.}
    \vspace{-0.25cm}
    \label{fig:mugbingham}
\end{figure}

The Bingham distribution, however, has a difficulty in calculating its normalizing constants, which is a key bottleneck of Bingham NLL computation.
Since the normalizing constant depends on the distribution's parameter, we must compute the constant for each parameter during optimization.
To our best knowledge, one common solution is to prepare a pre-computed table of the constants so the repetitive complicated calculation during optimization can be avoided. However, it takes time to create the table. 

In this paper, we introduce a fast-computable and easy-to-implement NLL function of Bingham distribution, which based on a novel algorithm by Chen \etal \cite{ChenTanaka2021}.
We also examine the parametrization of the Bingham distribution.
Furthermore, we will show its application to the object pose estimation in \secref{section:experiments}, 
and show how easy to apply our method to existing pose estimator.

\section{RELATED WORKS}

\subsection{Continuous Rotation Representations}
\label{section:rotationrepresentations}

4-dimensional rotation representation is widely used.
In particular, quaternion is a popular representation.
It is utilized such as in PoseCNN \cite{xiang2018posecnn}, PoseNet \cite{posenet2015}, and 6D-VNet \cite{Wenbin2021}.
Another 4-dimensional representation is an axis-rotation representation, introduced such as in MapNet \cite{mapnet2018}.
Although these 4D representation are valid in some cases, it is known that every $d$-dimensional ($d<5$) rotation representation is ``discontinuous'' (in the sense of \cite{Zhou2019}), because the 3-dimensional real projective space $\R P^3 (\cong \SO(3))$
cannot be embedded in $\R^d$ unless $d \geq 5$ \cite{Davis1998EmbeddingsOR}.
Because these representations has some singular points that prevent the network from stable regression,
it is preferred that we use the continuous rotation representation in training neural networks.

Researchers have proposed
various high-dimension representations.
For example, 9D representation was proposed in \cite{levinson20neurips} by using the singular value decomposition (SVD). 10D representation was proposed in \cite{peretroukhin_so3_2020} using the 4-dimensional symmetric matrix, which can be used for parametrization of Bingham distribution.
We will examine other continuous paramterization in \secref{section:rotationalambiguity}.

\subsection{Expressions of Rotational Ambiguity}
\label{section:rotationalambiguity}

There are several ways to express the rotational ambiguity.
In \cite{manhardt2019explaining, Bui2020}, they tried capturing the ambiguity by using multiple quaternions and minimizing the special loss function.
In KOSNet \cite{Hashimoto2019}, they used 3 parameters for describing camera's rotation (elevation, azimuth, and rotation around optical axis), and employed Gaussian distribution for describing their ambiguity.
These two representations, however, are discontinuous as described in \secref{section:rotationrepresentations}, since their dimensions are both lesser than 5.

Another approach is to use Bingham distribution.
This distribution is easy to parameterize and has been utilized in pose estimation field.
It is used for the model distribution of a Bayesian filter to realize online pose estimation \cite{Srivatsan2017},
for visual self-localization \cite{deng2020deep},
for multiview fusion \cite{Riedel2016},
and for describing the pose ambiguity of objects with symmetric shape \cite{Gilitschenski2020}.
Moreover, 
there is a continuous parameterization of this distribution, 
as Peretroukhin \etal \cite{peretroukhin_so3_2020} gives an example of it.
We adopt Bingham distribution as our probabilistic rotation representation.

\subsection{Loss Functions for Bingham Representation}

An NLL loss function is the common choice and has advantages, as described in the Introduction.
When computing  Bingham loss,
the main barrier is the computation of normalizing constant.
One solution of this problem is to take the time to pre-compute the table of normalizing constant. For example, Kume \etal ~\cite{Kume2013saddlepoint} use the saddlepoint approximation technique to construct this table.
However, even if we use the pre-computed table, we also need to implement a smooth interpolation function as described in \cite{Gilitschenski2020} to compute a constant missing in the table and a derivative of each constant for backpropagation. 
Thus, it is desirable to use Bingham's negative log likelihood (NLL) loss without the pre-computed table. 

Instead of using the Bingham NLL loss, 
another loss function is defined in \cite{peretroukhin_so3_2020}.
They defined QCQP loss to prevent them from suffering this troublesome computations. While this shows moderate performance by calculating uni-modal Bingham loss implicitly, it is difficult to extend to more complicated model such as mixture models \cite{deng2020deep, Riedel2016}, because the normalizing constant is required to mix multiple Bingham distributions. In contrast, our Bingham loss can be easily extended to mixture models.

\section{Rotation Representations}

\subsection{Quaternion and Spatial Rotation}

\subsubsection{Quaternion}
We introduce symbols $i,j,k$ which satisfies the property:
\begin{equation}
    i^2=j^2=k^2=ijk=-1.
    \label{eq:imaginaryunits}
\end{equation}
A quaternion is an expression of the form:
\begin{equation}
    q = w + xi+yj+zk
    \label{eq:quaternion}
\end{equation}
where $w,x,y,z$ are real numbers. $i,j,k$ are called the imaginary units of the quaternion.
The set of quaternions forms a 4D vector space whose basis is $\{1,i,j,k\}$.
Therefore, we identify a quaternion $q$ defined in \eqref{eq:quaternion} with
\begin{equation}
    \qt = (w,x,y,z)^\top \in \R^4.
\end{equation}

\subsubsection{Product of Quaternions}
For any quaternion $q' = a + bi + cj + dk$, 
we can define a product of quaternions $q'q$ thanks to the rule \eqref{eq:imaginaryunits}.
The set of quaternions forms a \textit{group} by this multiplication.
$q'q$ can also be identified with an element of $\R^4$. We denote it $\qt'\odot \qt \in \R^4$. 
Note that $q'q \neq qq'$ in general. Since $\qt' \odot \qt$ is bilinear w.r.t. $\qt' $ and $ \qt$, we can define matrices $\OmegaL(\qt')$ and $\OmegaR(\qt)$ satisfying
\begin{equation}
    \qt' \odot \qt = \OmegaL(\qt') \qt = \OmegaR(\qt) \qt'.
\end{equation}
$\OmegaL(\qt')$ and $\OmegaR(\qt)$ can be written in closed form:
\begin{align}
    \OmegaL(\qt') &= \begin{pmatrix}
        a & -b & -c & -d \\
        b & a & -d & c \\
        c & d & a & -b \\
        d & -c & b & a
    \end{pmatrix}, \label{eq:omegal}\\
    \OmegaR(\qt) &= \begin{pmatrix}
        w & -x & -y & -z \\
        x & w & z & -y \\
        y & -z & w & x \\
        z & y & -x & w
    \end{pmatrix}. \label{eq:omegar}
\end{align}

\subsubsection{Conjugate, Norm, and Unit Quaternion}
The \textit{conjugate} of $\qt$ is defined by $\qtc = (w,-x,-y,-z)^\top$. 
In general, $\conj{\qt \odot \qt'} = \qtc' \odot \qtc$. 
In particular, if $\qt = \qt'$, then 
\begin{equation}
    \qt \odot \qtc = \qtc \odot \qt = w^2 + x^2 + y^2 + z^2.
\end{equation}
By definition \eqref{eq:omegal} and \eqref{eq:omegar}, we get
\begin{equation}
    \OmegaL(\qtc) = \OmegaL(\qt)^\top, \quad \OmegaR(\qtc) = \OmegaR(\qt)^\top.
    \label{eq:conjandtranspose}
\end{equation}

We define the norm of quaternion $\|\bm{q}\|$ as
\begin{equation}
    \|\bm{q}\| = \sqrt{\qt \odot \qtc} = \sqrt{\qtc \odot \qt}.
\end{equation}
We call $\bm{q}$ a \textit{unit quaternion} if $\|\bm{q}\| = 1$. For a unit quaternion, its inverse coincides with its conjugate: $ \qt^{-1} = \qtc$.
Using \eqref{eq:conjandtranspose}, we can see that $\OmegaL(\qt)$ and $\OmegaR(\qt)$ are both orthogonal:
\begin{equation}
    \OmegaL(\qt)^\top \OmegaL(\qt) = \OmegaR(\qt)^\top \OmegaR(\qt) = I_{4}
\end{equation}
where $\qt$ is any unit quaternion, and $I_{4}$ is the 4-dimensional identity matrix.

We denote the set of unit quaternions $\Sp^3$ because it is homeomorphic to a 3-sphere $\Sp^3$.

\subsubsection{Unit Quaternion and Spatial rotation}
It is well known that unit quaternions can represent the spatial rotation. 
A mapping $R: \Sp^3 \to \SO(3)$ defined below is in fact a group homomorphism:
\begin{align}
    R(\qt) = \left(\hspace{-1.5mm}
    \begin{array}{ccc}
        1-2y^{2}-2z^{2} & -2 w z+2 x y & 2 w y+2 x z \\
        2 w z+2 x y & 1-2x^{2}-2z^{2} & -2 w x+2 y z \\
        -2 w y+2 x z & 2 w x+2 y z & 1-2x^{2}-2y^{2}
    \end{array}\hspace{-1.5mm}\right)\hspace{-1mm}.
    \label{eq:quaternion2so3}
\end{align}
Crucially, antipodal unit quaternions represent the same rotation; namely, $R(-\qt) = R(\qt)$.

\subsection{Definition of Bingham Distribution and Its Properties}
\label{section:bingham}

The Bingham distribution \cite{bingham1974} is a probability distribution on the unit sphere $\Sp^{d-1} \subset \R^{d}$ with the property of antipodal symmetry, which is consistent with the quaternion's property. 
We set $d=4$ throughout of this paper because we only consider $\Sp^3$.
We define \textit{Bingham distribution} as follows.
\begin{equation}
\bingham(\qt; \D, \eigvals) = \frac{1}{\normconst(\eigvals)} \exp\left( \qt^\top \D \diag(\eigvals) \D^\top \qt \right). \label{eq:binghamdefinition}
\end{equation}
where $\qt \in \Sp^3, \D\in \Ortho(4), \eigvals \in \R^4$.
Here $\Ortho(n)$ denotes the $n$-dimensional orthogonal group. Note that it is a Lie group and its Lie algebra $\ortho(n)$ is formed by $n$-dimensional skew-symmetric matrices.

Here we define $\diag : \R^m \to \R^{m\times m}$ as below.
\begin{equation}
    \diag : \begin{pmatrix}
        v_1 \\ \vdots \\ v_m
    \end{pmatrix}
    \mapsto \begin{pmatrix}
        v_1 & & \\
         & \ddots & \\
         & & v_m
    \end{pmatrix}
\end{equation}
$\normconst(\eigvals)$ is called a \textit{normalizing factor} or a \textit{normalizing constant} of a Bingham distribution $\bingham(\qt; \D, \eigvals)$.
$\normconst(\eigvals)$ is defined as below:
\begin{equation}
    \normconst(\eigvals) = \int_{\qt \in \Sp^3} \exp\left( \qt^\top \diag(\eigvals) \qt \right) \d_{\Sp^3} (\qt)
\end{equation}
where $\d_{\Sp^3}(\cdot)$ is the uniform measure on the $\Sp^3$.
Note that a normalizing factor depends only on $\eigvals$. 
It is easy to check that for any $c \in \R$, 
\begin{equation}
    \bingham(\qt ; D, \eigvals + c) = \bingham(\qt ; D, \eigvals)
\end{equation}
where $\eigvals + c = (\lambda_1 + c, \dots, \lambda_4 + c)$. Therefore, we can set $\eigvals$ satisfying
\begin{equation}
    0 = \lambda_1 \geq \lambda_2 \geq \lambda_3 \geq \lambda_4 \label{eq:sortedlambda}
\end{equation}
by sorting a column of $D$ if necessary.
A processed $D$ and a processed $\eigvals$ are denoted as $D_\text{shifted}$, $\eigvals_\text{shifted}$ respectively
It follows directly from the Rayleigh's quotient formula that
\begin{equation}
    \arg \max_{\qt \in \Sp^3} \bingham(\qt ; D, \eigvals) = \qt_{\lambda_1}
    \label{eq:modequaternion}
\end{equation}
where $\qt_{\lambda_1}$ is a column vector of $D$ corresponding to the maximum entry of $\eigvals$. If we sort $\eigvals$ as \eqref{eq:sortedlambda}, $\qt_{\lambda_1}$ coincides with the left-most column vector of $D$.

\subsection{Parametrization of Bingham Distribution}
\subsubsection{Representaion using Symmetric Matrix}

There are several choices of the parametrization of Bingham distribution. 
Firstly, we introduce here the 10D parameterization using a symmetric matrix which proposed by \cite{peretroukhin_so3_2020}.
Since every symmetric matrices can be diagonalized by some orthogonal matrix, we can rewrite the distribution instead of \eqref{eq:binghamdefinition}:
\begin{equation}
    \bingham_{\text{sym}}(\qt; A) = \frac{1}{\normconst(\eigvals)} \exp\left( \qt^\top A \qt \right),
\end{equation}
where $A$ is a 4-dimensional symmetric matrix. 
If the the eigenvalues of $A$ is sorted and shifted so as to satisfy \eqref{eq:sortedlambda}, then we call it $A_\text{shifted}$.
We assume that all parameters are shifted.

Here we define a bijective map $\triu : \Sym_4 \to \R^{10}$ as below:
\begin{equation}
    \triu : \begin{pmatrix}
            \theta_{1} & \theta_{2} & \theta_{3} & \theta_{4} \\
            \theta_{2} & \theta_{5} & \theta_{6} & \theta_{7} \\
            \theta_{3} & \theta_{6} & \theta_{8} & \theta_{9} \\
            \theta_{4} & \theta_{7} & \theta_{9} & \theta_{10}
    \end{pmatrix}
    \mapsto \begin{pmatrix}
        \theta_1 \\ \vdots \\ \theta_{10}
    \end{pmatrix}
\end{equation}
where $\Sym_n$ denotes the set of $n$-dimensional symmetric matrices. We can use this for 10D parametrization of Bingham distribution $\Param_{10}$ as following:
\begin{equation}
    \Param_{10}: \R^{10} \ni \bm{\theta} = 
    \begin{pmatrix}
        \theta_1 \\ \vdots \\ \theta_{10}
    \end{pmatrix}
     \mapsto \bingham_{\text{sym}}\left(\qt; \triu(\bm{\theta}) \right).
\end{equation}
We call this representation \textit{Peretroukhin representation} here.

\subsubsection{Representations of Orthogonal Matrices}

The Peretroukhin representation defined above is simple; however, it includes the eigenvalue decomposition process for calculate the normalizing factor $\normconst(\eigvals)$, which has a high computational cost.
It is reasonable that the network directly infers a orthogonal matrix $D$ and a diagonal entries $\eigvals$, then reconstructs $A = D\diag(\eigvals)D^\top$.

To parametrize $D \in \Ortho(4)$, we introduce following two strategies.
The first is Cayley transformation \cite{Hairer2006}, which is commonly used representation for orthogonal matrices.
It is expressed as follows:
\begin{equation}
    \cayley: \R^6 \ni \bm{\theta} \mapsto \left(I-S(\bm{\theta})\right)^{-1} \left(I+S(\bm{\theta})\right) \in \Ortho(4)
\end{equation}
where $S: \R^6 \to \ortho(4)$ is defined as following:
\begin{equation}
    S:
    \begin{pmatrix}
        \theta_1 \\ \vdots \\ \theta_{6}
    \end{pmatrix}
    \mapsto
    \begin{pmatrix}
        0 & \theta_1 & -\theta_2 & \theta_3 \\
        -\theta_1 & 0 & \theta_4 & -\theta_5 \\
        \theta_2 & -\theta_4 & 0 & \theta_6 \\
        -\theta_3 & \theta_5 & -\theta_6 & 0
    \end{pmatrix}.
\end{equation}
This representation can be express all of the orthogonal matrices.
We call this orthogonal matrix representation, $\cayley$, a \textit{Cayley representation}.

The second is to use 4-dimensional representation defined by Birdal et al.\cite{birdal2018}.
Nevertheless there is a orthogonal matrix that cannot be represented in 4D, it works well in such as \cite{deng2020deep}.
This representation uses the orthogonal property of unit quaternion's matrix representation. That is, as in \eqref{eq:omegal},
\begin{equation}
    \birdal :\R^4 \ni \bm{\theta} =
    \begin{pmatrix}
        \theta_1 \\ \vdots \\ \theta_{4}
    \end{pmatrix}
    \mapsto
    \OmegaL(\bm{\theta}) \in \Ortho(4).
\end{equation}
We call this orthogonal matrix representation $\birdal$ a \textit{Birdal representation}.

\subsubsection{Representations of Eigenvalues}

The simplest choice for representation of $\eigvals$ is that the network infers $\eigvals \in \R^4$ directly, then shift and sort it so as to satisfy \eqref{eq:sortedlambda}:
\begin{equation}
    \Lambda_4:
    \R^4 \ni 
    \begin{pmatrix}
        \theta_1 \\ \theta_2 \\ \theta_3 \\ \theta_{4}
    \end{pmatrix}
    \mapsto 
    \begin{pmatrix}
        0 \\ \lambda_2 - \lambda_1 \\ \lambda_3 - \lambda_1 \\ \lambda_4 - \lambda_1
    \end{pmatrix} \in \R^4
    \label{equ:repr_lambda4}
\end{equation}
where $\lambda_i$ $(i=1,\dots,4)$ is a permutation of $\{\theta_1,\dots,\theta_4\}$ satisfying $\lambda_1 \geq \dots \geq \lambda_4$.

Another approach is to use softplus function \cite{softplus2010} defined as below:
\begin{equation}
    \phi(x) = \log(1+\exp(x)).
\end{equation}
Note that $\phi(x) > 0$ for all $x\in \R$.
Using the softmax function, we can define a 3D representation as follows \cite{deng2020deep}:
\begin{equation}
    \Lambda_3: \R^3 \ni 
    \begin{pmatrix}
        \theta_1 \\ \theta_2 \\ \theta_3
    \end{pmatrix}
    \mapsto
    \begin{pmatrix}
        0 \\ -\phi(\theta_1) \\ -\phi(\theta_1)-\phi(\theta_2) \\ -\phi(\theta_1)-\phi(\theta_2)-\phi(\theta_3)
    \end{pmatrix} \in \R^4
    \label{equ:repr_lambda3}
\end{equation}
The resulting tuple automatically satisfies \eqref{eq:sortedlambda}. 

So far we introduced two representations for $D$ and two for $\eigvals$. 
Now we have 5 choices of parametrization of Bingham distribution:
\begin{equation}
\begin{array}{crr@{\,\,\mapsto\,\,}l}
    \bullet& \Param_{10}:& \R^{10} \ni \bm{\theta} & \bingham_{\text{sym}}\left(\qt; \triu(\bm{\theta}) \right)\\
    \bullet& \Param_{4+3}:& \R^4\times \R^3 \ni (\bm{d}, \eigvals) & \bingham\left(\qt; \birdal(\bm{d}), \Lambda_3 (\eigvals) \right)\\
    \bullet& \Param_{4+4}:& \R^4\times \R^4 \ni (\bm{d}, \eigvals) & \bingham\left(\qt; \birdal(\bm{d}), \Lambda_4 (\eigvals) \right)\\
    \bullet& \Param_{6+3}:& \R^6\times \R^3 \ni (\bm{d}, \eigvals) & \bingham\left(\qt; \cayley(\bm{d}), \Lambda_3 (\eigvals) \right) \\
    \bullet& \Param_{6+4}:& \R^6\times \R^4 \ni (\bm{d}, \eigvals) & \bingham\left(\qt; \cayley(\bm{d}), \Lambda_4 (\eigvals) \right)
    \label{eq:listofparams}
\end{array}
\end{equation}

Note that these representations are all continuous in the sense of \cite{Zhou2019}.
We will compare these representations in \secref{section:experiments}.

\begin{algorithm}[tb]
   \caption{Our implementation of the loss function}
   \label{algo:lossfunction}
   \begin{algorithmic}[1]
   
   \Function {Integrator}{$f_\text{integrant}$, $\eigvals$}
       \State $N_\text{min} \gets 15$; $N \gets 200$
       \State $r\gets 2.5$; $\omega_d \gets 0.5$
       \State Define $c$ as in \eqref{eq:vars}; $d \gets c / 2$
       \State Define $h,p_1,p_2$ as in \eqref{eq:vars}
       \State $S\gets 0$
        \For {$n = -N-1,\dots,N$}
        \State $S \gets S + w(|nh|)\cdot f_\text{integrant}(nh, \eigvals)\cdot e^{nh\sqrt{-1}}$ 
        \Statex \Comment{$w$ is defined in \eqref{eq:weightfunc}}
        \EndFor
       \State \Return the real part of $\pi e^c h S$
   \EndFunction
   \State
   \Function {BinghamLoss}{$D$, $\eigvals$, $\qt_\text{gt}$}
        \State $D_\text{shifted}$, $\eigvals_\text{shifted}$ $\gets$ \Call{Sort\&Shift}{$D$, $\eigvals$}
        \State $A_\text{shifted}$ $\gets$ $D_\text{shifted} \diag(\eigvals_\text{shifted}) D_\text{shifted}^\top$
        \State $\normconst \gets $ \Call{Integrator}{$\integrand$, $\eigvals$} \Comment{see \eqref{eq:def_C}}
        \State \Return $-\qt_\text{gt}^\top A_\text{shifted}\, \qt_\text{gt} + \ln \normconst$
    \EndFunction

   \end{algorithmic}
\end{algorithm}

\section{EFFICIENT COMPUTATION OF LOSS FUNCTION}
\label{section:computationofloss}
\subsection{Definition of Loss Function}
The negative log-likelihood function of the Bingham distribution can be written as follows:
\begin{equation}
    \mathcal{L}(D, \eigvals, \qt_\text{gt}) = -\qt_\text{gt}^\top D \diag(\eigvals) D^\top \,\qt_\text{gt} + \ln \normconst(\eigvals).
\end{equation}
It had been a hard problem to compute $\normconst(\eigvals)$ until a high efficient computation method was proposed by \cite{ChenTanaka2021}.
Our loss function is implemented mainly based on \cite{ChenTanaka2021}.

\subsection{Calculation of Normalizing Constant and Its Derivative}

The whole procedure is shown in \algoref{algo:lossfunction}.
Let $r, \omega_d$ be real numbers satisfying 
\begin{equation}
    r\geq 2 \quad\text{and}\quad \frac{1}{r} \leq \omega_d \leq 1.
\end{equation}
We chose $r = 2.5,\,\omega_d = 0.5$ here.
Let $c,h,p_1,p_2$ be defined as
\begin{equation}
    \begin{array}{c}
    \displaystyle
    c = \frac{N_\text{min} \pi }{r^2(1+r) \omega_d},\quad h = \sqrt{\frac{2\pi d (1+r)} {\omega_d N}}, \\[12pt]
    \displaystyle
    p_1 = \sqrt{\frac{Nh}{\omega_d}},\quad p_2 = \sqrt{\frac{\omega_d N h}{4}},
    \end{array}
    \label{eq:vars}
\end{equation}
where $d$ is any positive number satisfying $d < c$. We chose $d = c/2$ here.
$N$ is a positive integer satisfying $N \geq N_\text{min}$.
One can choose $N_\text{min}$ arbitrarily; however, a too small $N_\text{min}$ may lead to unstable computation. 
We chose $N_\text{min} = 15$ here.

We define a function $w$ parametrized by $p_1,\,p_2$ in \eqref{eq:vars} as below.
\begin{equation}
    w(x) = \frac{1}{2} \erfc \left( \frac{x}{p_1} - p_2 \right),
    \label{eq:weightfunc}
\end{equation}
where $\erfc$ is the complementary error function:
\begin{equation}
    \erfc(x) = 1-\frac{2}{\sqrt{\pi}} \int_{0}^{x} e^{-t^{2}} d t.
\end{equation}
Then we define
\begin{align}
    \integrand (t, \eigvals) &= \prod^4_{k=1} \left(-\lambda_k + t\sqrt{-1} + c\right)^{-1/2}, \\
\intertext{and, for each $i = 1,\dots, 4$, we get}
    \frac{\partial \integrand}{\partial \lambda_i}(t, \eigvals) &= \frac{1}{2}\left( -\lambda_i + t\sqrt{-1} + c \right)^{-1}\integrand(t, \eigvals).
\end{align}

Now we can calculate the normalizing constant $\normconst$ as below
\begin{align}
    \normconst(\eigvals) &= \pi e^c h \sum_{n=-N-1}^N w(|nh|)\, \integrand(nh, \eigvals)\, e^{nh\sqrt{-1}}, \label{eq:def_C}\\
    \frac{\partial \normconst}{\partial \lambda_i}(\eigvals) &= \pi e^c h \sum_{n=-N-1}^N w(|nh|)\, \frac{\partial \integrand}{\partial \lambda_i}(nh, \eigvals)\, e^{nh\sqrt{-1}}, \label{eq:def_dCdLam}
\end{align}
for each $i = 1,\dots, 4$.
Although a calculation result of $\normconst(\eigvals)$ and ${\partial \normconst}/{\partial \lambda_i}(\eigvals)$ should exactly be a real number, one may get a complex number with the very small imaginary part.
In our implementation shown in \algoref{algo:lossfunction}, we ignore the imaginary part, assuming that it is sufficiently small.

It is noteworthy that if we set the true value of $\normconst(\eigvals)$ as $\normconst_\text{truth}(\eigvals)$, we get
\begin{equation}
    |\normconst_\text{truth}(\eigvals) - \normconst(\eigvals)| = O\left(\sqrt{N} e^{-c\sqrt{N}}\right)
\end{equation}
for a constant $c>0$ independent from $N$ \cite{TANAKA201473}. This means that we can achieve any accuracy if we set a large enough $N$.
In this paper, we set $N = 200$ in consideration of the computation time.

\section{APPLICATION TO POSE ESTIMATOR}
\label{section:experiments}

\begin{figure}[t]
    \centering
    \includegraphics[width=\linewidth]{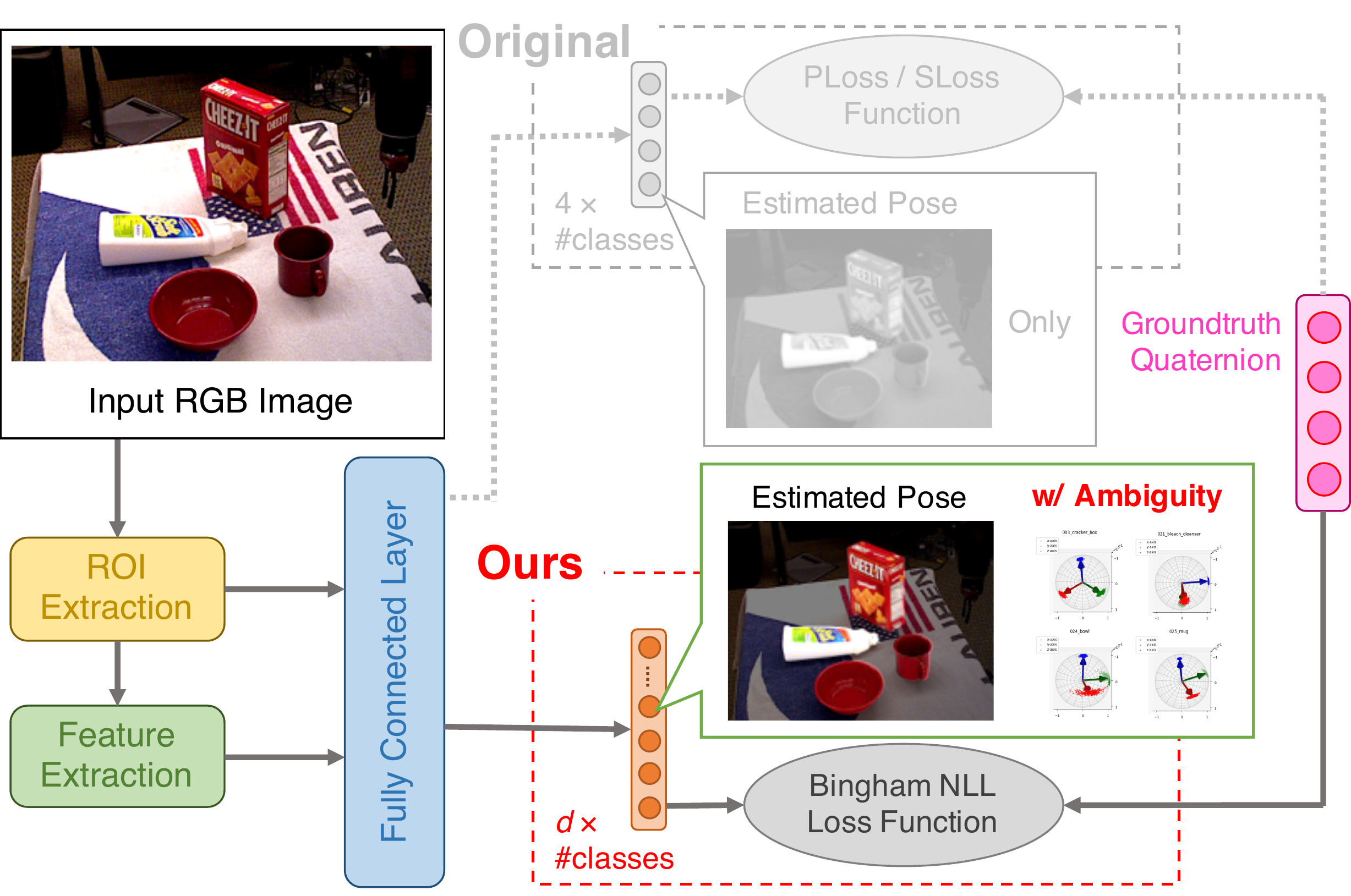}
    \caption{Overview of our network implementation. We changed the loss function and the dimension of final output from 4 to $d$.
    As we will describe in \secref{section:experimentsresults}, we decided $d=10$.
    Our method can also obtain the pose ambiguity.
    }
    \label{fig:bingham_posecnn}
\end{figure}

\subsection{Implementation}
In this section, we introduce an application of our representation to an existing pose estimator.
We use PoseCNN \cite{xiang2018posecnn} as the backborn framework.
Our implementation is based on PoseCNN-PyTorch \cite{posecnnpytorch} by 
NVIDIA Research Projects. 
We call our network ``\textit{Bingham-PoseCNN}'' for convenience.
\figref{fig:bingham_posecnn} shows the overview of Bingham-PoseCNN.
``Ours'' in the figure is the point of change compared to the original PoseCNN.
We changed just the dimension of the final FCN layer from 4 to $d$
($d$ varies from 7 to 10).

On the one hand, our Bingham NLL loss function is described in \secref{section:computationofloss}.
On the other hand, original PoseCNN used the following loss functions for training;
\begin{align}
    \operatorname{PLOSS}(\qt, \qt_\text{gt})&=\frac{1}{2 m} \sum_{\bm{x} \in \mathcal{M}}\|R(\qt_\text{gt}) \bm{x}-R(\qt) \bm{x}\|^{2}, \\
    \operatorname{SLOSS}(\qt, \qt_\text{gt})&=\frac{1}{2 m} \sum_{\bm{x}_{1} \in \mathcal{M}} \min _{\bm{x}_{2} \in \mathcal{M}}\left\|R(\qt_\text{gt}) \bm{x}_{1}-R(\qt) \bm{x}_{2}\right\|^{2},
\end{align}
where $\mathcal{M}$ denotes the set of points on the mesh model of each object.
They used PLoss if the object has no symmetry, and SLoss if the object has symmetry.
While they annotated the symmetric property to each object, our method doesn't need these annotations.
In addition, our method doesn't need mesh models of target objects.

\subsection{Dataset}

We tested our model with the YCB-Video dataset which is the same dataset used in \cite{xiang2018posecnn}.
In this dataset, 80 videos for training, and 2949 keyframes for testing that are extracted from the rest 12 unused videos were provided.
In addition, YCB-Video dataset contains 80000 synthetic images. We also used them for training.

\subsection{Evaluation Metrics}

We used ADD and ADD-S metrics to evaluate the performance of our pose estimator.
Let $(R,\bm{t})$ be a pair of the groundtruth rotation $R$ and translation $\bm{t}$, and $(\widehat{R},\widehat{\bm{t}})$ be a pair of the estimated rotation and translation. 
Then, 
ADD and ADD-S are defined as below:
\begin{align}
    \text{ADD}&= \frac{1}{m}\sum_{\bm{x} \in \mathcal{M}} \|(R\bm{x} + \bm{t}) - (\widehat{R}\bm{x} + \widehat{\bm{t}})\|,\\
    \text{ADD-S}&= \frac{1}{m}\sum_{\bm{x}_1 \in \mathcal{M}} \min_{\bm{x}_{2} \in \mathcal{M}} \|(R\bm{x}_1 + \bm{t}) - (\widehat{R}\bm{x}_2 + \widehat{\bm{t}})\|,
\end{align}
where $\mathcal{M}$ is the set of the $m$ sampled points from 3D mesh model's surface.
These metrics are the same as that used in PoseCNN \cite{xiang2018posecnn}.

\subsection{Results}
\label{section:experimentsresults}
\subsubsection{Comparison of Bingham Representations}

\tabref{tab:comparisonParam} shows the mean values of area under the curve (AUC) of ADD and ADD-S.
Peretroukhin's 10D representation achieves the best score among 6 representations.
Thus we adapt this representation to our network.
8D representation with Birdal representation comes next.

It can be seen that the 4D eigenvalue representations
got a higher score than 3D representation
.
This would be related that $\Lambda_4$, defined in \eqref{equ:repr_lambda4}, tends to give smaller eigenvalues than the values given by $\Lambda_3$, defined in \eqref{equ:repr_lambda3}.
It is known that the distribution is concentrated if the eigenvalues are small, as we will described in \secref{section:epismeticuncertainty}.
In the well-trained network, the mode quaternion closer to the given quaternion as the dispersion of the distribution becomes smaller.
This is a possible reason why $\Lambda_4$ gives a better result that $\Lambda_3$.

\begin{table}[t]
    \centering
    \caption{Comparison of Parametrization. Only the mode quaternion is used for ADDs calculation for Bingham representation.}
    \label{tab:comparisonParam}
    \begin{tabular}{c|c|c|r||c|c}
    \hline
    Param & \multicolumn{1}{l|}{Ortho. Matrix} & Diag. Entries & Dim & ADD & ADD-S \\ \hline
    $\vphantom{\frac{1}{2}}$$\Param_{4+3}$ & \multirow{2}{*}{Birdal \cite{birdal2018}} & 3D            & 7 & 49.2 & 72.9      \\ \cline{1-1}\cline{3-6}
    $\vphantom{\frac{1}{2}}$$\Param_{4+4}$ &                         & 4D            & 8 & 53.0 & 74.2      \\ \hline
    $\vphantom{\frac{1}{2}}$$\Param_{6+3}$ & \multirow{2}{*}{Cayley \cite{Hairer2006}} & 3D            & 9 & 13.5 & 58.8     \\ \cline{1-1}\cline{3-6}
    $\vphantom{\frac{1}{2}}$$\Param_{6+4}$ &                 & 4D            & 10 & 23.5 & 66.2     \\ \hline
    $\vphantom{\frac{1}{2}}$$\Param_{10}$ & \multicolumn{2}{c|}{Symmetry Matrix \cite{peretroukhin_so3_2020}}               & 10 & \textbf{55.1} & \textbf{75.1}     \\ \hline \hline
    -- & \multicolumn{2}{c|}{Quaternion \cite{xiang2018posecnn}}                    & 4 & 52.9 & 74.1    \\ \hline
    \end{tabular}
\end{table}

\subsubsection{Conventional Quaternion vs Bingham's Mode Quaternion}

\begin{table*}[t]
    \centering
    \caption{Area under the curve of \figref{fig:results}. ``Ratio'' is the ratio of Ours score to Original PoseCNN's score. Ratio $>$ 95.0\% are shown in bold.}
    \label{tab:addandadds}
    \begin{tabular}{l|cc|cc|cc||cc|cc|cc}
        \hline
        & \multicolumn{6}{c||}{RGB}                                                                      & \multicolumn{6}{c}{RGB + Depth}                                                              \\ \hline
        & \multicolumn{2}{c|}{ADD} & \multicolumn{2}{c|}{ADD-S} & \multicolumn{2}{c||}{Ratio} & \multicolumn{2}{c|}{ADD} & \multicolumn{2}{c|}{ADD-S} & \multicolumn{2}{c}{Ratio} \\
        \multicolumn{1}{c|}{objects} & Ours & Original & Ours & Original & ADD            & ADD-S          & Ours & Original & Ours & Original & ADD            & ADD-S          \\ \hline
        002\_master\_chef\_can   & 61.0       & 60.5       & 86.7        & 88.7        & \textbf{100.9}     & \textbf{97.7}     & 70.7       & 70.0       & 93.1        & 93.4        & \textbf{101.0}     & \textbf{99.7}     \\
        003\_cracker\_box        & 36.6       & 61.2       & 63.3        & 79.5        & 59.8               & 79.6              & 64.8       & 79.1       & 72.8        & 85.4        & 81.9               & 85.3              \\
        004\_sugar\_box          & 56.7       & 51.6       & 76.7        & 73.4        & \textbf{109.8}     & \textbf{104.5}    & 91.3       & 90.7       & 94.7        & 93.7        & \textbf{100.6}     & \textbf{101.1}    \\
        005\_tomato\_soup\_can   & 71.1       & 69.5       & 83.7        & 82.6        & \textbf{102.4}     & \textbf{101.3}    & 86.8       & 87.8       & 93.2        & 93.5        & \textbf{98.8}      & \textbf{99.7}     \\
        006\_mustard\_bottle     & 88.7       & 84.5       & 94.0        & 92.1        & \textbf{105.0}     & \textbf{102.1}    & 94.9       & 90.6       & 96.6        & 93.5        & \textbf{104.8}     & \textbf{103.2}    \\
        007\_tuna\_fish\_can     & 73.5       & 68.4       & 91.5        & 87.8        & \textbf{107.5}     & \textbf{104.2}    & 84.7       & 84.3       & 96.9        & 95.1        & \textbf{100.4}     & \textbf{101.9}    \\
        008\_pudding\_box        & 29.3       & 67.8       & 53.8        & 83.4        & 43.2               & 64.5              & 81.8       & 86.0       & 90.3        & 93.7        & \textbf{95.1}      & \textbf{96.4}     \\
        009\_gelatin\_box        & 87.8       & 80.2       & 92.8        & 89.4        & \textbf{109.4}     & \textbf{103.8}    & 65.4       & 95.3       & 67.4        & 97.2        & 68.7               & 69.4              \\
        010\_potted\_meat\_can   & 60.1       & 59.7       & 79.3        & 78.2        & \textbf{100.6}     & \textbf{101.3}    & 80.4       & 78.6       & 89.4        & 88.3        & \textbf{102.3}     & \textbf{101.2}    \\
        011\_banana              & 69.8       & 77.4       & 85.4        & 89.9        & 90.2               & \textbf{95.0}     & 81.2       & 89.1       & 90.0        & 95.1        & 91.2               & 94.6              \\
        019\_pitcher\_base       & 68.6       & 67.8       & 84.4        & 83.8        & \textbf{101.1}     & \textbf{100.6}    & 92.3       & 93.9       & 96.2        & 96.5        & \textbf{98.3}      & \textbf{99.7}     \\
        021\_bleach\_cleanser    & 50.7       & 51.1       & 67.5        & 70.1        & \textbf{99.2}      & \textbf{96.3}     & 85.2       & 84.2       & 93.9        & 91.5        & \textbf{101.2}     & \textbf{102.6}    \\
        024\_bowl                & 4.3        & 4.9        & 60.5        & 74.2        & 87.7               & 81.6              & 17.9       & 8.4        & 90.4        & 78.1        & \textbf{213.8}     & \textbf{115.7}    \\
        025\_mug                 & 71.2       & 47.4       & 87.8        & 72.4        & \textbf{150.3}     & \textbf{121.4}    & 77.7       & 84.6       & 88.8        & 95.0        & 91.9               & 93.5              \\
        035\_power\_drill        & 61.4       & 52.7       & 77.9        & 72.7        & \textbf{116.6}     & \textbf{107.2}    & 87.4       & 86.3       & 92.8        & 91.7        & \textbf{101.4}     & \textbf{101.2}    \\
        036\_wood\_block         & 0.9        & 1.3        & 21.7        & 15.8        & 69.7               & \textbf{137.3}    & 39.5       & 29.0       & 86.2        & 88.7        & \textbf{136.1}     & \textbf{97.1}     \\
        037\_scissors            & 43.5       & 50.1       & 65.1        & 68.8        & 86.7               & 94.6              & 62.1       & 72.8       & 77.1        & 82.2        & 85.3               & 93.8              \\
        040\_large\_marker       & 55.1       & 55.2       & 66.9        & 67.2        & \textbf{99.8}      & \textbf{99.5}     & 82.3       & 86.1       & 90.4        & 93.2        & \textbf{95.6}      & \textbf{96.9}     \\
        051\_large\_clamp        & 43.2       & 12.9       & 68.4        & 38.9        & \textbf{334.9}     & \textbf{175.8}    & 63.1       & 56.7       & 81.6        & 76.8        & \textbf{111.3}     & \textbf{106.3}    \\
        052\_extra\_large\_clamp & 8.1        & 6.3        & 37.6        & 38.6        & \textbf{128.5}     & \textbf{97.5}     & 27.3       & 9.2        & 49.6        & 40.6        & \textbf{295.8}     & \textbf{122.1}    \\
        061\_foam\_brick         & 50.3       & 56.8       & 83.9        & 90.0        & 88.5               & 93.2              & 63.4       & 67.2       & 95.0        & 96.6        & 94.3               & \textbf{98.3}     \\ \hline
        all                      & 55.1       & 53.0       & 75.1        & 74.1        & \textbf{104.0}     & \textbf{101.4}    & 75.5       & 75.8       & 88.3        & 88.7        & \textbf{99.6}      & \textbf{99.5}    \\ \hline
            \end{tabular}
\end{table*}

\tabref{tab:addandadds} shows the result of PoseCNN with our Bingham representation and conventional quaternion representation.
Mode quaternion described in \eqref{eq:modequaternion} is used for comparing the performance of Bingham representation with that of conventional one.
Bingham representation with our loss function achieved a equivalent performance with that of quaternions.

\section{DISCUSSIONS}

\subsection{Evaluation of Inferred Probabilistic Representation}

\tabref{tab:addandadds} is evaluated only with the mode quaternion.
Our method can extract information about the ambiguity or uncertainty of inferrence result.
Inferred results are probability distribution so we may interpret them in a several way. We evaluate the result in the two interpretation: confidence and shape ambiguity.

\subsubsection{Interpret as Confidence (Epismetic uncertainty)}
\label{section:epismeticuncertainty}
\figref{fig:mugbingham} shows an example of inferrence result of our model.
The plots in the right column shows sampled poses from the inferred distribution. The sampling algorithm from Bingham distribution is based on 
\cite{Kent2013BinghamSampling}.

In \figref{fig:mugbingham},
If the handle of the mug appears, the resulting distribution becomes low-variance and concentrated. This can be interpreted that the inferred result has high confidence.
In contrast, if the handle is occluded, the distribution becomes widely spread. This can be interpreted that the inferred result has low confidence.

To explain this more quantitatively, 
here we introduce a random variable $\DeltaQ$.
Given a random quaternion from the estimated distribution $Q \sim \bingham(D,\eigvals)$ and the groundtruth $\qt_\text{gt}$, we define a random variable $\DeltaQ$ as follows:
\begin{align}
    \DeltaQ &= 2\arccos\left( \left| \Re (Q \odot \qt_\text{gt}^{-1}) \right| \right) \nonumber \\
    &= 2\arccos \left( \left|{\qt_\text{gt}^\top Q}\right| \right),
    \label{eq:deltaQ}
\end{align}
where $\Re(\qt)$ is the real part of $\qt$; that is, if $\qt = w + xi + yj + zk$, then $\Re(\qt) = w$.
A realization of $\DeltaQ$ represents a difference between sampled rotations from the inferred distribution and the groundtruth.

\begin{figure}[t]
    \centering
    \includegraphics[width=0.8\linewidth]{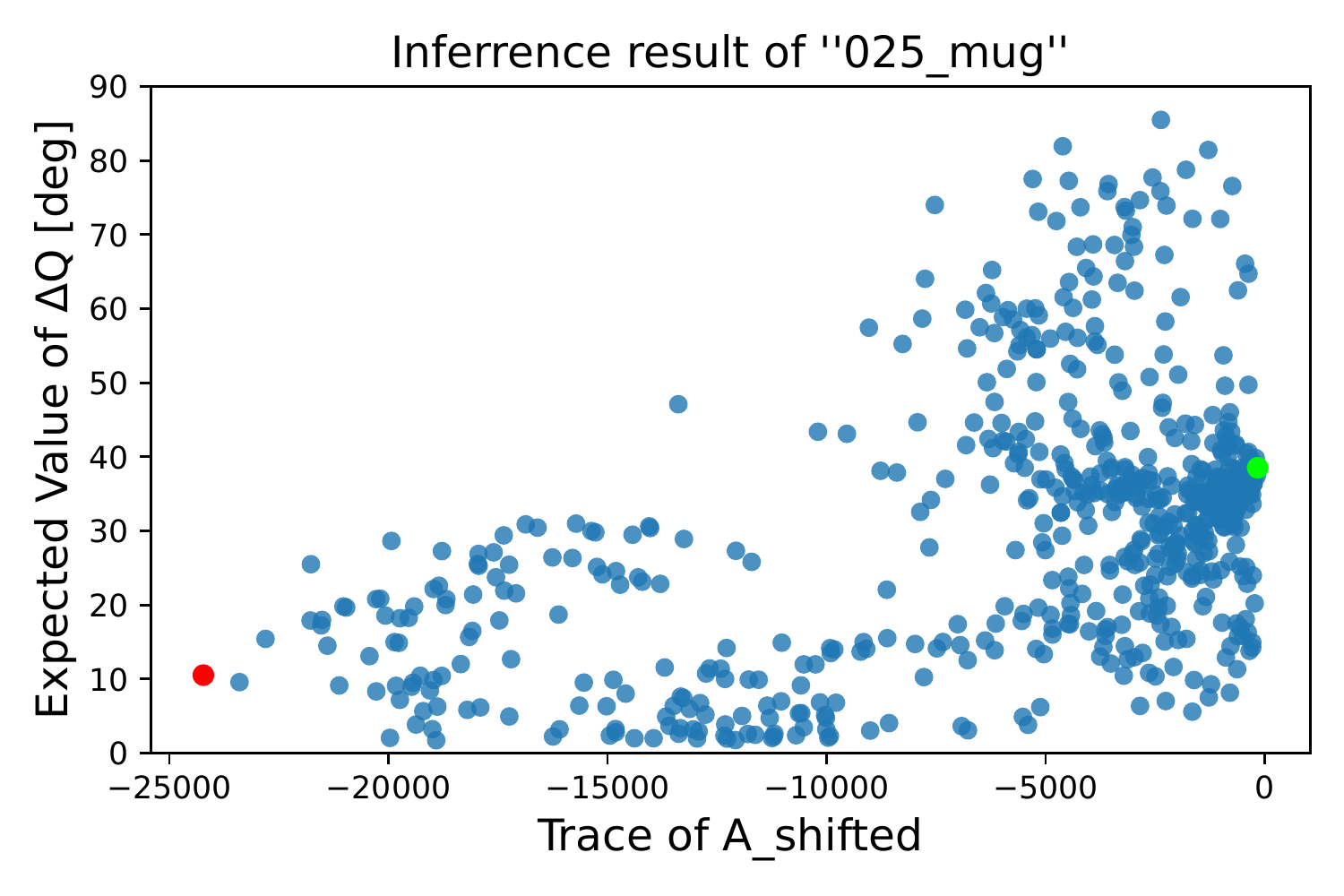} \\
    \includegraphics[width=\linewidth]{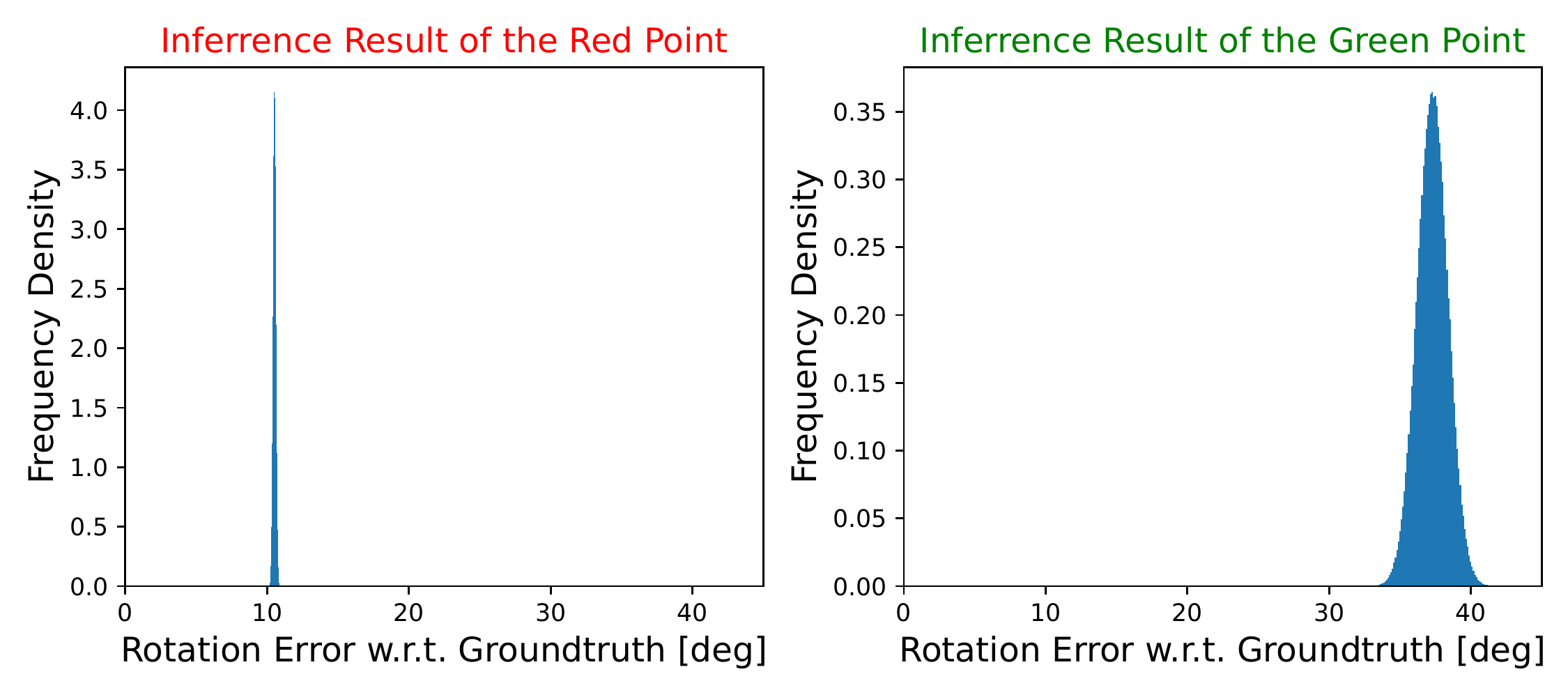}
    \caption{
    An example of the inference result of our network. The upper figure shows the relation between $\Expect{\DeltaQ}$ and $\trace(A_\text{shifted})$. The red and the green point are the minimum and the maximum trace, respectively.
    The lower figures are the distribution of $\DeltaQ$ at the red and the green point shown in the upper figure.
    }
    \label{fig:evaluationofprob}
\end{figure}

Now we introduce an indicator of the inference uncertainty proposed in \cite{peretroukhin_so3_2020}.
They empirically found that as the trace of shifted parameter matrix: the lesser
\begin{equation}
    \trace(A_\text{shifted}) = \trace(A) - 4 \max (\eigvals)
\end{equation}
becomes, the more confident the estimation is.
The upper figure in \figref{fig:evaluationofprob} shows the inferrence result of \texttt{025\_mug} as an example. The red and green plots in the figure are corresponding to the minimum and maximum value of traces, respectively. 
The lower figures 
shows the distribution at the corresponding point in the upper figure; the left is to the red point, and the right is to the green point, respectively.
Here we can see that if the trace is large, then $\DeltaQ$ widely distributes, that is, the confidence is low.
In addition, we can also see that $\Expect{\DeltaQ}$ becomes smaller as the trace is lesser. This implies that an inference with large trace may have a large error.

\subsubsection{Interpret as Rotation Symmetry (Aleatoric uncertainty)}

In \figref{fig:mugbingham}, we can see that rotations are zonally spread around the $z$-axis. We can interpret this that the mug in this view has rotational symmetry around the z-axis.
We can see the symmetry characteristics of the observed objects quantitatively 
by inspecting the eigenvalues.
According to \cite{Kunze2004InterpletLambda}, 
for the eigenvalues sorted as \eqref{eq:sortedlambda}, $\eigvals$ gives
\begin{itemize}
    \item a bipolar distribution, if $\lambda_2 + \lambda_3 < \lambda_4$,
    \item a circular distribution, if $\lambda_2 + \lambda_3 = \lambda_4$,
    \item a spherical distribution, if $\lambda_2 + \lambda_3 > \lambda_4$,
    \item a uniform distribution, if $\lambda_2 = \lambda_3 = \lambda_4$.
\end{itemize}
The orientation of the symmetry axis is determined by the orthogonal matrix $D$.

\begin{figure}[t]
    \centering
    \includegraphics[width=\linewidth]{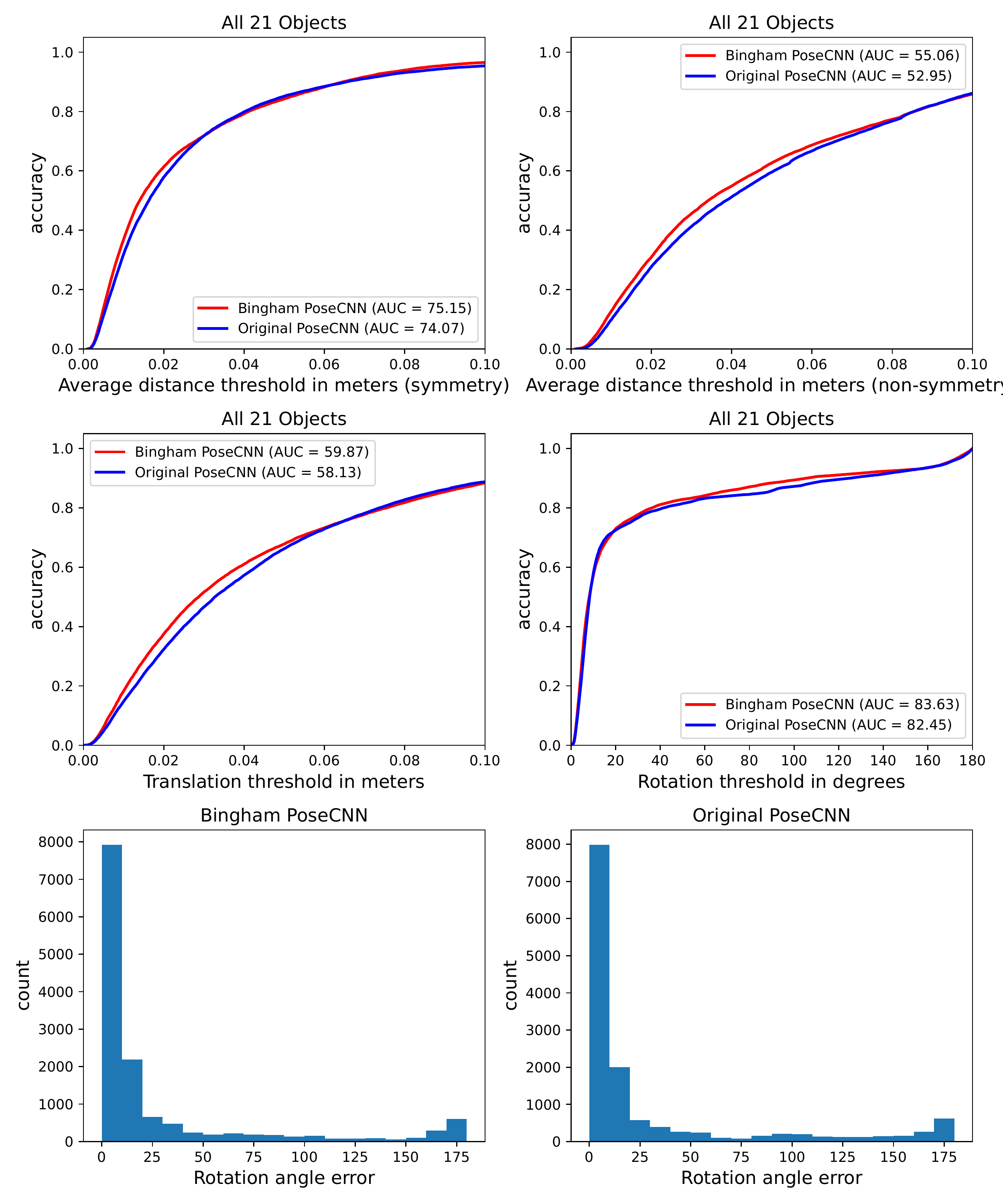}
    \caption{Results on ours and PoseCNN \cite{xiang2018posecnn}: Top row represents Average distance threshold curves. Middle row represents Translation threshold curves. Bottom row represents the histogram of rotation angle error. These metrics are described in \cite{xiang2018posecnn}.}
    \label{fig:results}
  \end{figure}

\subsection{Explanation How Our Bingham Representation Works}

\begin{figure}[t]
    \centering
    \includegraphics[width=\linewidth]{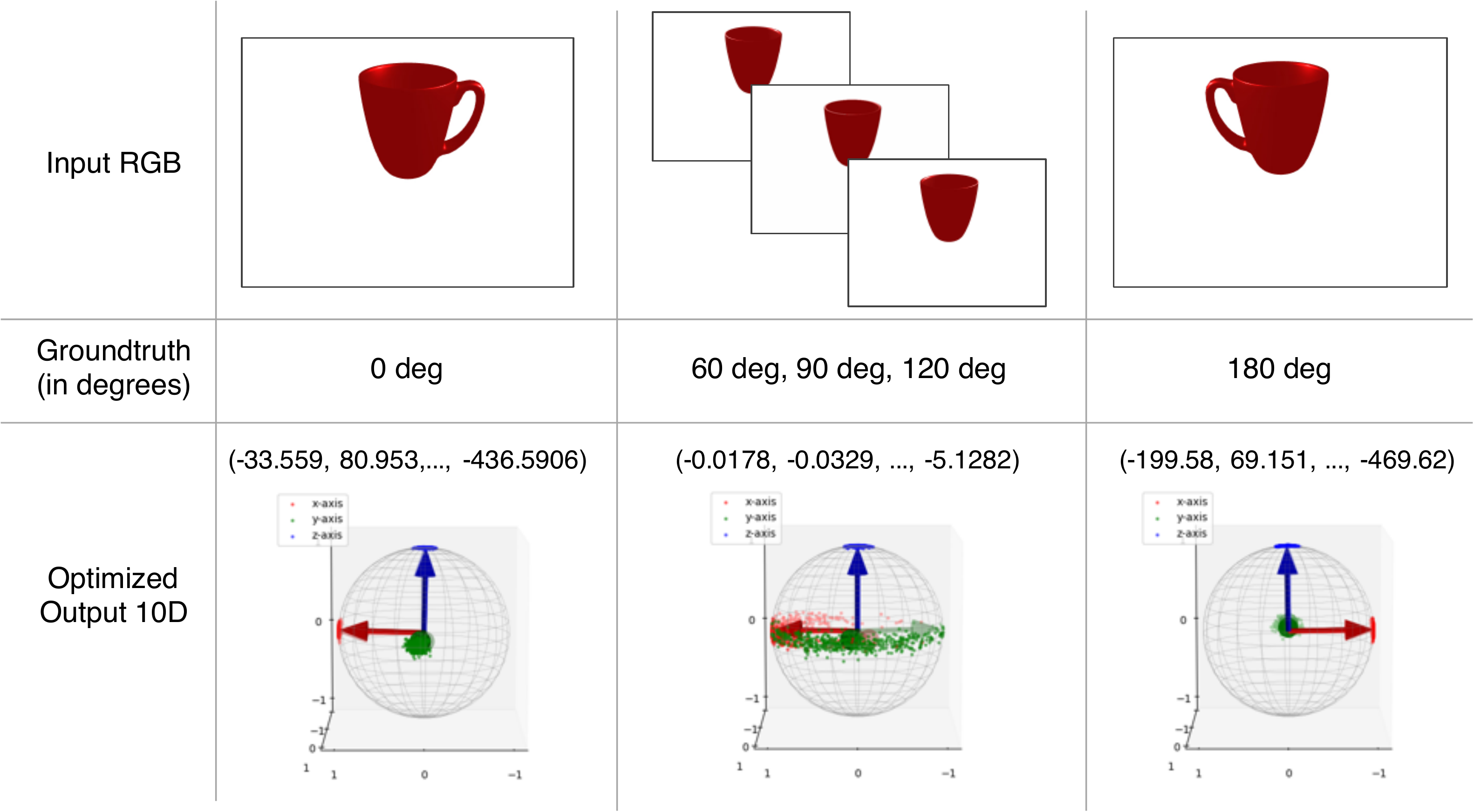}
    \caption{Some toy example for an explanation of the mechanism how the network learns the Bingham parameter. Groundtruths are shown in rotation angles around the $z$-axis instead of quaternions. The first two and the last entries of inferred 10D parameter are shown in the bottom row.}
    \vspace{-0.6cm}
    \label{fig:mechanismofinferrence}
  \end{figure}

\figref{fig:mechanismofinferrence} shows synthetic images of a red cup.
It shows the groundtruth of the angles of rotation around $z$-axis, instead of the quaternions.
The inferred distributions with their 10D parameters are also shown.
We will use this figure for explaining the learning mechanism of our network.

Our model is mathematically represented as below:
\begin{equation}
    \bm{\theta}_i = F(I_i) 
\end{equation}
where $I_i$ is an input image and $\bm{\theta}_i$ is a inferred parameter of distribution. $F$ is our network to be trained.
Suppose that we have pairs $\{I_i,\qt_i\}_{i=1}^N$ whose images are similar to each other but whose groundtruth quaternions are all different.
In \figref{fig:mechanismofinferrence}, the middle column is corresponding to this circumstance.
In this situation, there is a $\bm{\theta}$ satisfying
\begin{equation}
    F(I_i) \approx \bm{\theta} \quad \text{for all $i = 1,\dots,N$}.
\end{equation}
Let $M$ be a function that transforms a given parameter vector to a parameter matrix of Bingham distribution.
Then our problem becomes
\begin{center}
    Find $\bm{\theta}$ that minimize $\displaystyle \sum_{i=1}^N \mathcal{L}(M(\bm{\theta}), \qt_i)$.
\end{center}
By solving this problem, we finally get $\bm{\theta}$ which is optimized to the given all quaternions $\qt$.
The problem is equivalent to ``solve the maximum likelihood estimation (MLE) problem for each $M(f(I_i))$, given $\qt_i$''.
This means that the resulting parameter has information about the distribution of quaternions that share the similar views.

In \figref{fig:mechanismofinferrence}, the inferred results in the left and right column are both concentrated because quaternions that gives the similar view are close to each other. In contrast, the result in the middle column is zonally spread because the quaternions sharing the view is widely spread. Our network learns the parameter that covers rotations sharing the similar view.

\subsection{Adapting to Objects with Discrete Symmetry}

In \tabref{tab:addandadds}, the objects with discrete symmetry, such as  \texttt{036\_wood\_block} and \texttt{052\_extra\_large\_clamp}, got relatively low score in ADD and ADD-S.
This is because a single Bingham distribution cannot capture the ambiguity with multiple modes well.
We can improve score by introducing mixture Bingham representation which is introduced in such as \cite{deng2020deep}.
Our NLL loss function is easy to extend to them, compared to non-NLL losses such as the QCQP loss presented in \cite{peretroukhin_so3_2020}.

\section{CONCLUSIONS}

We proposed and implemented a Bingham NLL loss function which is free from pre-computed lookup table.
This is directly computable and there is no need to interpolate computation.
Also, we showed our loss function is easy to implement 
for being used for training.
Moreover, it is quite easy to be introduced to the existing 6D pose estimator. 
We tested with PoseCNN as an example and proved that our representation successfully expressed the ambiguity of rotation while the evaluating the peak of distribution showed equivalent performance with that of original PoseCNN.
Furthermore, we discovered the relationship between the various parametrization of the Bingham distribution and the performances from object pose perspective.
In future works, we would like to handle mixture Bingham distribution for more capabilities, especially for the objects with discrete symmetry, based on this loss function.








\bibliographystyle{IEEEtran}
\bibliography{IEEEabrv,myrefs}
\end{document}